\documentclass[10pt,journal,compsoc]{IEEEtran}
\usepackage{amsmath,amsfonts}
\usepackage{array}
\usepackage[caption=false,font=normalsize,labelfont=sf,textfont=sf]{subfig}
\usepackage{textcomp}
\usepackage{stfloats}
\usepackage{url}
\usepackage{verbatim}
\usepackage{graphicx}
\usepackage{color}
\usepackage{multirow}
\usepackage{comment}
\usepackage[linesnumbered,ruled,lined,commentsnumbered]{algorithm2e}
\ifCLASSOPTIONcompsoc
  \usepackage[nocompress]{cite}
\else
  \usepackage{cite}
\fi
\ifCLASSINFOpdf
\else
\fi
\begin{document}
%
\title{Uncertainty-oriented Order Learning for Facial Beauty Prediction}
%
%
%
%

\author{Xuefeng Liang,~\IEEEmembership{Member,~IEEE,} Zhenyou Liu, Jian Lin, Xiaohui Yang and Takatsune Kumada
\IEEEcompsocitemizethanks{\IEEEcompsocthanksitem X. Liang (corresponding author) and X. Yang are with the School of Artificial Intelligence, Xidian University, Xi’an 710071, China\protect\\
E-mail: xliang@xidian.edu.cn
\IEEEcompsocthanksitem J. Lin and Z. Liu are with the Guangzhou Institute of Technology, Xidian University, Guangzhou, 510555, China 
\IEEEcompsocthanksitem T. Kumada is with the Graduate School of Informatics, Kyoto University, Kyoto, Japan.
\IEEEcompsocthanksitem Jian Lin and Zhenyou Liu contributed equally to this work.}}

%
%

\markboth{Journal of \LaTeX\ Class Files,~Vol.~14, No.~8, August~2015}%
{Shell \MakeLowercase{\textit{et al.}}: Bare Demo of IEEEtran.cls for Computer Society Journals}
%



\IEEEtitleabstractindextext{%
\begin{abstract}
Previous Facial Beauty Prediction (FBP) methods generally model FB feature of an image as a point on the latent space, and learn a mapping from the point to a precise score. Although existing regression methods perform well on a single dataset, they are inclined to be sensitive to test data and have weak generalization ability. 
We think they underestimate two inconsistencies existing in the FBP problem: 1. inconsistency of FB standards among multiple datasets, and 2. inconsistency of human cognition on FB of an image. 
To address these issues, we propose a new Uncertainty-oriented Order Learning (UOL), where the order learning addresses the inconsistency of FB standards by learning the FB order relations among face images rather than a mapping, and the uncertainty modeling represents the inconsistency in human cognition. 
The key contribution of UOL is a designed distribution comparison module, which enables conventional order learning to learn the order of uncertain data. 
Extensive experiments on five datasets show that UOL outperforms the state-of-the-art methods on both accuracy and generalization ability.
\end{abstract}

}

\maketitle

\IEEEdisplaynontitleabstractindextext

%
\IEEEpeerreviewmaketitle

\IEEEraisesectionheading{\section{Introduction}\label{sec:introduction}}

%
%
%
%
\IEEEPARstart{S}{ociological} and psychological studies \cite{ref21} have shown that Facial Beauty (FB) has a great impact on career development, interpersonal relationships, social status and social acceptance. 
Thus, Facial Beauty Prediction (FBP), a challenging task in computer vision, has attracted much attention. 
In the last decade, several methods had been applied for FBP \cite{ref3,ref31,ref42}, automatic facial beautification \cite{ref27}, and makeup recommendation \cite{ref2,ref32,ref37,ref40}.
The pioneer FBP methods focused on designing handcrafted features based on aesthetic knowledge, such as geometric features \cite{ref1,ref10,ref12,ref17,ref34}, holistic features \cite{ref13,ref35,ref45,ref46,ref50,ref54} or mixed features \cite{ref3,ref9,ref19,ref49,ref58}, and performing prediction by classifiers (such as KNN \cite{ref1,ref45,ref9}, SVM \cite{ref34,ref49,ref5,ref51}, decision trees \cite{ref34}, Adaboost \cite{ref45}, etc.) or regression algorithms (such as linear regression \cite{ref10,ref19,ref41}, ridge regression \cite{ref35,ref50}, Gaussian regression \cite{ref51}, etc.). 

\begin{figure}[t]
	\centering
	\includegraphics[width=0.8\linewidth]{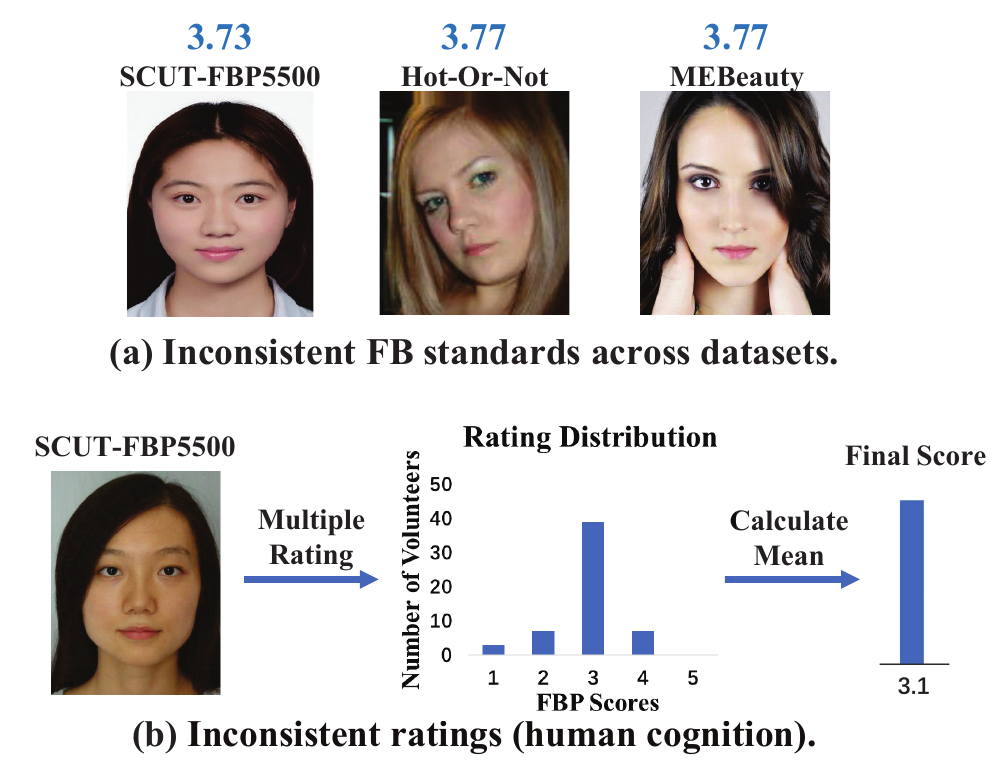}
	\setlength{\abovecaptionskip}{0.1cm}
	\caption{Two inconsistencies in FBP problem. (a). Three images, coming from SCUT-FBP5500, Hot-Or-Not and MEBeauty datasets respectively, have similar normalized FB scores but different FB appearances. (b). Ratings of a face image from different people are commonly inconsistent. Many FBP methods take the mean of these ratings as the FB score.}
	\vspace{-15pt}
	\label{fig:incons}
\end{figure}

Later, deep learning was introduced into FBP due to its superiority at various vision tasks. Many studies tried to learn the mapping from FB features to FB scores (the mean of multiple ratings) \cite{ref31,ref42,ref30,ref52,ref53}. 
Although existing regression methods perform well on a single dataset, they often show weak generalization when tested across datasets. 
We think they underestimate two inconsistencies existing in the FBP: \IEEEpubidadjcol (1) \emph{Inconsistency of FB standards.} For different FB datasets, the volunteers who rate the facial images are often different populations with different cultural and educational backgrounds, and thus have different reference systems for their FB standards.
Even FB scores across datasets are normalized to the same scale, biases still exist, as shown in Fig. \ref{fig:incons} (a). In our experiment, many volunteers ranked the right face in the MEBeauty dataset as the most beautiful one, and the left face in the SCUT-FBP5500 as the second one; (2) \emph{Inconsistency of human cognition.} 
Studies \cite{ref11,ref48} pointed that the FB ratings made by different people were more likely to diverge. Figure \ref{fig:incons} (b) illustrates an example.

The inconsistency of FB standards among datasets has not been addressed in this field yet, but could be regarded as a nonlinear label shift in domain adaptation problem, in which different datasets are the overlapping subsets of the universal domain. 
Existing domain adaptation methods aim to learn domain invariant representations from multiple datasets via minimizing domain shift measures \cite{ref59}, optimal distribution matching \cite{ref60,ref61} and domain adversarial training \cite{ref62}. On the contrary, we expect to learn the invariant in FB from a single dataset, which can be easily applied to other datasets. 
Our observation shows the order of FB is highly consistent across different datasets and can be learned from one FBP dataset. 
A psychology study \cite{ref44} has shown that human subconsciously cognize real-world scales by learning an order rather than measuring exact values. 
An order pattern is essentially an awareness of relative relation that is independent on the reference base. 
The research \cite{ref39} also pointed out that relative relations can be measured much easier than estimating precise quantities in many cases.
Lin et al. \cite{ref30} introduced the relative ranking into the loss function of a regression model for FBP problem.
Lim et al. \cite{ref29} proposed an order learning based on relative relations, and applied it to estimate precise facial ages, which demonstrated a better performance. 
We then apply the idea of order learning to learn the FB order of instances in the dataset to address the problem of the inconsistency of FB standards.

For the inconsistency of human cognition, some studies \cite{ref11,ref48} applied the label distribution learning (LDL) as the objective of regression model. 
However, LDL essentially learns the mapping from a feature point on the latent space to a label distribution, and is still inclined to overfit the data. 
In psychophysics, Thurstone proposed \emph{A Law of Comparative Judgment} \cite{ref36} to address such inconsistency, which is also known as uncertainty problem. 
Thurstone argued that the discriminal processes generated by a stimulus were not always equal, therefore, modeled it as a Gaussian distribution on a psychological scale, known as \textbf{discriminal dispersion}.
Inspired by this, we model the inconsistent cognition of FB as a multi-dimensional Gaussian distribution on a high-dimensional psychological scale space.

However, conventional order learning can only compare data with precise labels rather than uncertain data. In order to address both inconsistencies, we design a module to compare distributions based on the Monte Carlo sampling, which enables order learning to learn the order relations of uncertain data.

Moreover, to compare with competing methods, the order relations must be transferred to FB scores.
To this end, order learning needs a reference set that must be balanced, continuous, and cover entire range. 
Unfortunately, FBP datasets are usually small, unbalanced (i.e. medium ratings are majority), even discontinuous. 
To relax this constraint, we introduce the Bradley-Terry model \cite{ref6}. 
It applies the maximum likelihood to estimate overall distribution using partial comparison results.

We conduct extensive experiments on the FBP benchmark dataset SCUT-FBP5500 and related datasets, Color FERET, Hot-Or-Not, MEBeauty and MIFS. The results show that our method outperforms six competing methods.

The main contributions of our work are threefold:

$\bullet$ We propose Uncertainty-oriented Order Learning (UOL), which enables order learning to learn the relative relations of uncertain data by a distribution comparison module.

$\bullet$ To address the inconsistency of FB standards among datasets, we apply order learning to learn the relative relations between instances. To address the inconsistency of human cognition, we model FB features as multi-dimensional Gaussian distributions on a psychological scale space, which can learn more robust relative relations of FB features. 

$\bullet$ Extensive experiments demonstrate that our UOL has a better performance and generalization over the competing methods on SCUT-FBP5500 and other FBP datasets.

\section{Related Work}
\subsection{Facial Beauty Prediction} 
The earliest methods of FBP generally used handcrafted features, such as geometric features \cite{ref1,ref10,ref12,ref17,ref34}, holistic features (e.g., color features \cite{ref50,ref19}, eigenface \cite{ref3,ref35,ref45,ref9}, LBP features \cite{ref13,ref46,ref54}, etc.) or mixed features (i.e., geometric and holistic features), then built classification \cite{ref1,ref34,ref9,ref49,ref5} or regression \cite{ref10,ref35,ref50,ref19,ref51,ref42} models.

Recently, convolutional neural networks (CNN) have become a mainstream method for FBP. 
Some studies considered FBP as a classification task. 
Gan et al. \cite{ref14} proposed the 2MBeautyNet to improve the accuracy. 
Zhai et al. developed three models consecutively, i.e. the BeautyNet model \cite{ref55}, the AFBS model \cite{ref56} and the BLS method \cite{ref57}. 

More studies treated FBP as a regression problem. 
Xu et al. \cite{ref52} proposed a CRNet that jointly optimized the classification branch and regression branch through classification and regression losses. Later, they introduced an improved expectation loss into ComboLoss \cite{ref53} to boost the performance. 
Lin et al. \cite{ref30} proposed a $R^3$CNN that used the relative ranking based loss. 
Shi et al. \cite{ref42} used pixels to mark different parts of a face as meta information and applied the co-attention learning mechanism to characterize the importance of different regions and different facial components simultaneously. 
Lin et al. \cite{ref31} proposed a facial attribute aware convolution neural network, AaNet, which used a parameter generator to adaptively adjust the filter of the main network.
F. Bougourzi et al. \cite{ref63} aimed to train a better regression model, so they proposed CNN-ER and applied dynamic loss parameters to minimize the effect of outliers.

As most FBP datasets are scored by multiple people, an alternative solution for FBP is label distribution learning (LDL). 
Fan et al. \cite{ref11} used LDL to train CNN on the basis of residual neural network. 
Wang et al. \cite{ref48} proposed the LDL method LDL-LDM to exploit global and local label correlations on label distribution manifolds.

\begin{figure*}[t]
	\centering
	\includegraphics[width=0.95\linewidth]{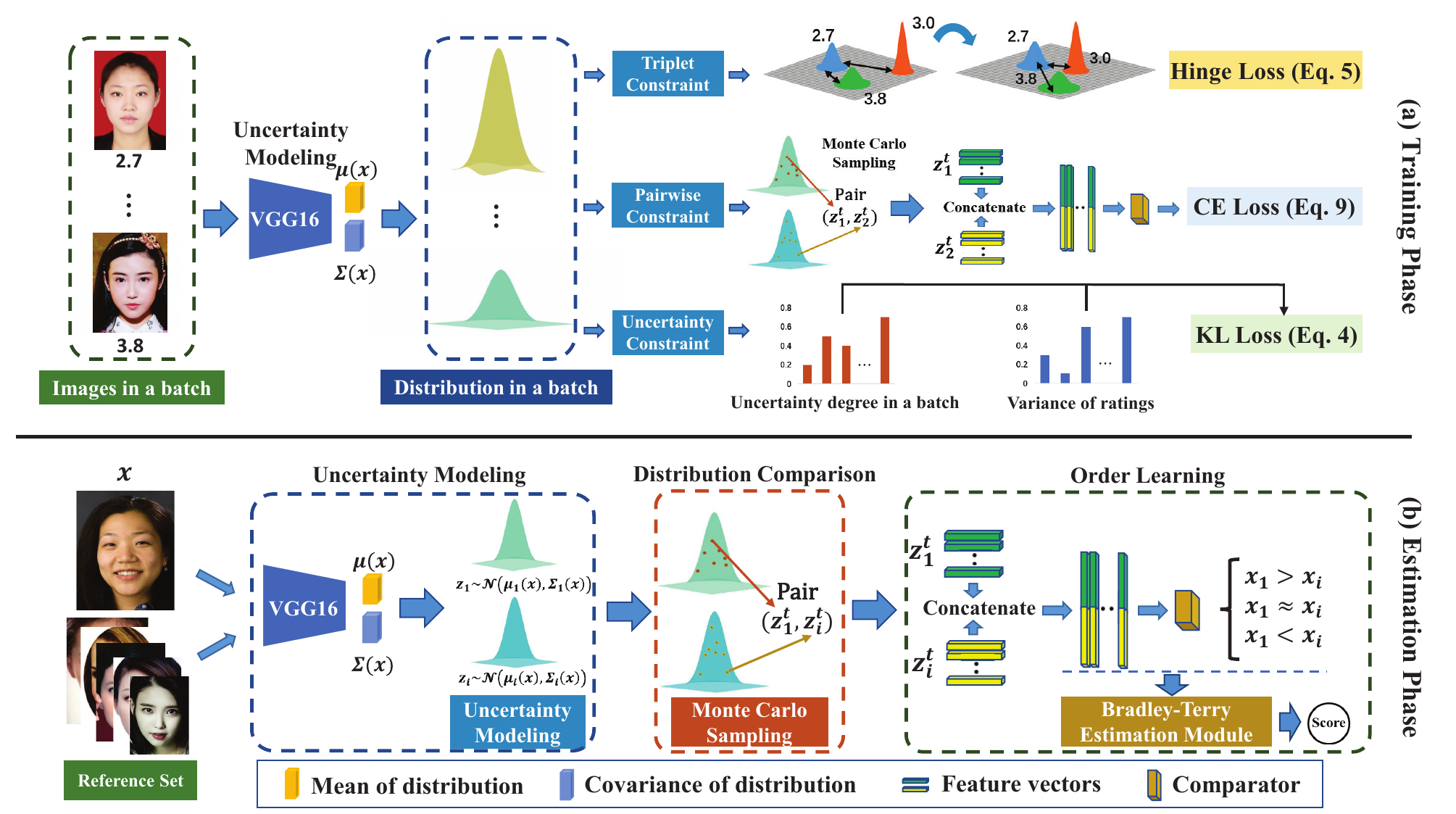}
	\setlength{\belowcaptionskip}{0.1cm}
	\caption{(a) The training phase of UOL. The order of distributions is constrained by cross entropy loss and hinge loss, and the dispersion of the distributions is constrained by KL loss. (b) The estimation phase of UOL. In uncertainty modeling, the FB of a facial image is modeled by a multi-dimensional Gaussian distribution whose mean $\mu$ and diagonal covariance $\Sigma$ are learned by VGG from the image. In distribution comparison, we sample from both the distributions of test image and reference image to form a pair and predict its order by a comparator in order learning. After having the order relations of $T$ pairs between reference images and the test image, the Bradley-Terry model is applied to estimate the score of the test image.}
	\vspace{-0.5cm}
	\label{fig:framework}
\end{figure*}

All above methods model the FB features of a facial image as a point on the latent space, and learn a mapping between the point and the given FB score, ranking or label distribution. Few of them consider the human cognition bias and the FB standard bias across datasets. Such direct regression mappings may suffer from weak generalization. 

\subsection{Order Learning and Pairwise Comparison} 
In everyday life, people often cognize the world and make decisions through comparisons. 
In psychology, kinds of abstract attitude are commonly measured by pairwise comparison, in which the subject compares a series of objects in pairs and makes a choice between two objects based on a certain criterion. 
Some studies have introduced pairwise comparison into their tasks. 
Ko et al. \cite{ref20} proposed a Pairwise Aesthetic Comparison Network (PACNet) to extract features for image aesthetic assessment (IAA). 
Lee et al. \cite{ref23} established the prototype of order learning, constructed a pairwise comparison matrix to predict image aesthetic scores. 
Hu et al. \cite{ref18} proposed a learning framework based on pairwise comparison by focusing on the relative quality ranking of restored images. 
Lim et al. \cite{ref29} formally proposed order learning and applied it to facial age estimation. 
It aims to discover the order of sequential patterns for a robust classification task. 
Both pairwise comparison methods and order learning can only work on data with certain label other than uncertain data, as they need to construct pairs for comparison. 

\subsection{Thurstone’s Theory} 
Thurstone argued that a term is needed for the process by which the organism identifies, distinguishes, discriminates, or reacts to stimuli. It is known as the discriminal process and also called as attitude. Psychologically some of these attributes can be measured. FBP is a classical attitude measurement in the field of psychophysics, and the inconsistency of human cognition has been studied by psychologists for a long time. Thurstone’s study showed that the discriminal processes generated by a stimulus were subject to noticeable fluctuation due to the different subjects who perceive the stimulus or the different environments. This fluctuation among the discriminal processes for a uniform repeated stimulus was designated the \textbf{discriminal dispersion}.

In his study, experiments showed that the discriminal dispersion which any given repeated stimulus produces on the psychological continuum is usually Gaussian. 
So Thurstone proposed the \emph{Law of Comparative Judgment} \cite{ref36} based on this theory and used the means and variances of discriminal dispersions to measure the attitudes on psychological scale. 
Later, Cohen \cite{ref64} extended Thurstone's theory to multi-dimensional spaces.

\subsection{Uncertainty Modeling}
The uncertainty reflects the dispersion of a random variable. Since the FB ratings encode human cognition bias, FB is a kind of uncertain data. 
There have been a few preliminary works to model the uncertainty for other tasks. 
Gast et al. \cite{ref15} proposed Probout to replace the intermediate activations with low-dimensional Gaussian distributions by adjusting the activation function, and obtained uncertainty in a lightweight manner instead of traditional Bayesian approaches. 
Liu et al. \cite{ref33} considered the image quality as a distribution rather than a feature point, then modeled the uncertainty by a low-dimensional Gaussian distribution. 
However, low-dimensional Gaussian distribution naturally limited the feature expressiveness.
In the field of face recognition, to reduce the uncertainty caused by image distortion, Shi et al. \cite{ref43} applied probabilistic face embeddings to model the uncertainty of face features. 
Chang et al. \cite{ref7} proposed data uncertainty learning (DUL) to achieve a joint learning of data embeddings and uncertainty. 
In the field of age estimation, Li et al. \cite{ref25} proposed a probabilistic ordinal embedding (POEs) to treat each facial age data as a multivariate Gaussian distribution and applied a set of ordinal distribution to enforce ordinal property in the embedding space. 
Although DUL and POEs modeled facial images as multi-dimensional Gaussian distributions, this uncertainty aimed to alleviate the effect of the inherent noise in the image, so only the ``mean" of distribution is used for inference. Also, to avoid the distribution degenerating into deterministic embedding, the information bottleneck loss in POEs only ensures that the covariance matrix of distribution is close to the Identity matrix $I$. 
Instead, the uncertainty in our UOL denotes the inconsistency of human cognition, so we expect that the FB distribution to be close to the \emph{discriminal dispersion} of human rating, and then take the variance of multiple ratings as the uncertainty to constrain the distribution.
Thus, both the motivation and modeling approach of UOL have differences from above methods.

\section{Methods}
Our proposed UOL consists of four modules: an order learning model in \textbf{section \ref{3.1}}; an uncertainty modeling module based on Thurstone’s \emph{discriminal dispersion} theory in \textbf{section \ref{3.2}}; a distribution comparison module in \textbf{section \ref{3.3}}, which enables order learning to learn the relative relations of uncertain data; and a FB score estimation module based on the Bradley-Terry model, which transforms the order relations to FB scores in \textbf{section \ref{3.5}}. Figure \ref{fig:framework} shows the overall framework.
\subsection{Order Learning}
\label{3.1}
Order learning aims to learn the order relations between instances. As the order between FB is independent on the reference base, it can largely avoid the bias of FB standards introduced by different datasets. Following is the principle of order learning.
Given two faces images, $x_i$ and $x_j$, and their FB scores $y_i$ and $y_j$ respectively, the order between $x_i$ and $x_j$ can be defined and encoded by a one-hot label,
\begin{align}
	Y=\begin{cases} x_i\approx x_j: [1,0,0],\ if \left\vert y_i-y_j\right\vert \le \theta,   \\ x_i<x_j: [0,1,0],\ if\  y_i-y_j<-\theta,\\ x_i>x_j: [0,0,1],\ if\  y_i-y_j>\theta ,\end{cases} 
\end{align}
where $\theta=0.2$ is the threshold that represents the discrimination of FB.

The conventional order learning treats an instance as a fixed feature point on the latent space, which is learned by a network $g(\cdot)$, shown in the left of Fig.\ref{fig:modeling}. It then carries out pairwise feature comparisons between two instances. 
The comparator $f (\cdot,\cdot)$ in order learning consists of three fully connected layers and is formulated as 
\begin{align}
	Y'= f(g(x_1),g(x_2)), 
\end{align}
which learns order relation from precise labels of two samples in the pair.

\subsection{Uncertainty Modeling}
\label{3.2}
\begin{figure}[t]
	\centering
	\includegraphics[width=0.999\linewidth]{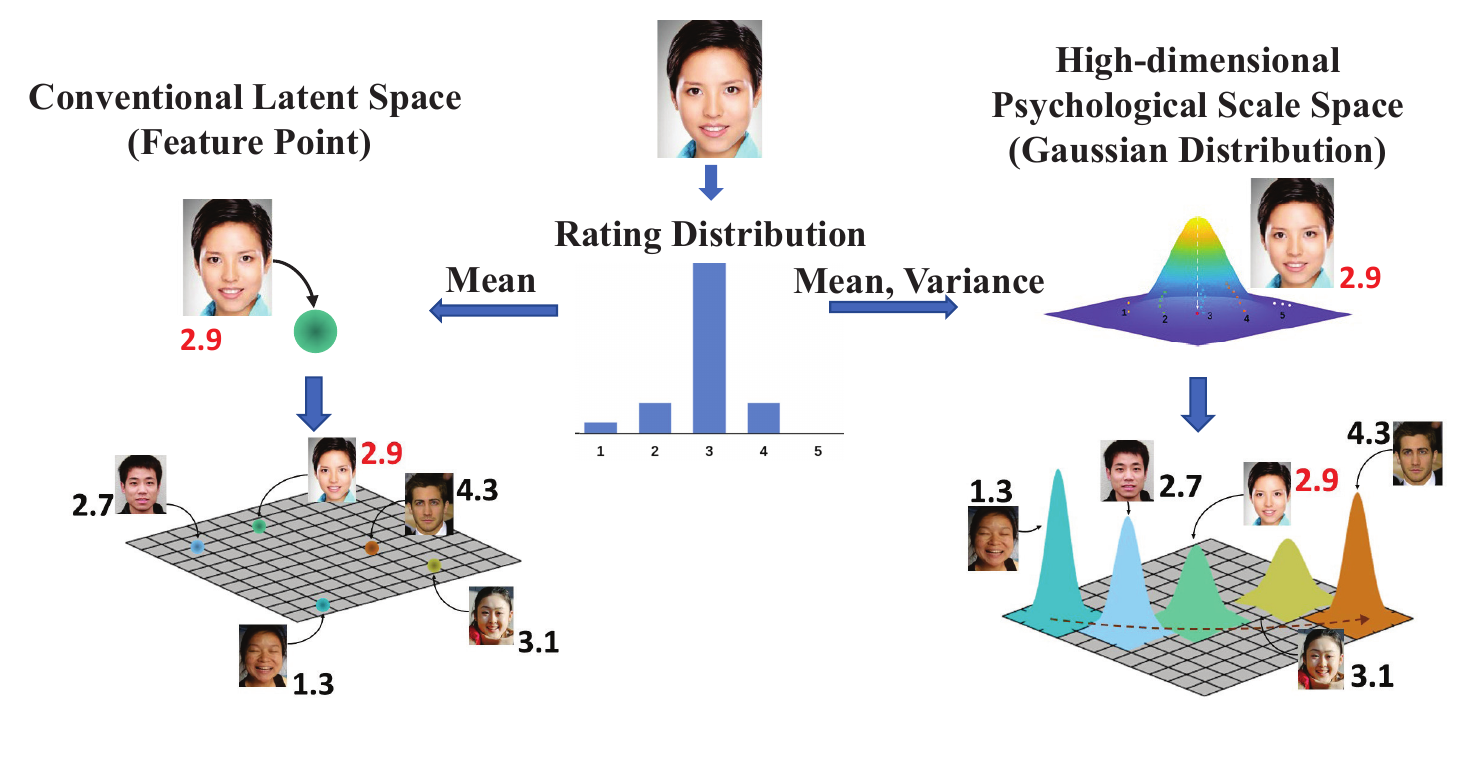}
	\caption{In our UOL, FB features of facial images are modeled as the multi-dimensional Gaussian distributions on psychological scale space instead of fixed points on the conventional latent space.}
	\label{fig:modeling}
\end{figure}
Previous methods of FBP usually use the mean of human ratings as the FB score for regression, which underestimate the human cognition biases. 
According to Thurstone’s \emph{discriminal dispersion} theory, the discriminal processes to the same stimulus are not always equal, but rather present a Gaussian distribution on the psychological scale. 
This theory is also applicable to FBP. 
We then design a high-dimensional psychological scale space to address the inconsistency of human cognition on FB. 
Specifically, we model the human ratings of an instance $x$ as a multi-dimensional Gaussian distribution $z \sim \mathcal{N}(\mu(x),\Sigma(x))$ in the space, which is used as a feature for pairwise comparisons, as shown in the right of Fig. \ref{fig:modeling}. 
A VGG16 is applied to encode mean vector $\mu(x)$ and covariance matrix $\Sigma(x)$ of the distribution. 
$\Sigma(x)$ represents the dispersion of the rating distribution. As $\Sigma(x)$ is a diagonal matrix, the degree of \emph{discriminal dispersion} is the Frobenius norm of it,\begin{align}
	\Vert \Sigma(x)\Vert _F = \sqrt{\sum_{i=j=1}^D {\vert \sigma _{i,j}\vert}}\ .
\end{align}

During training, the variance of the $i$-th instance’s ratings is set as the ground truth, $\eta_i$, representing \emph{discriminal dispersion} degree of the instance, and the network is optimized by minimizing the KL divergence between the predicted distribution of dispersion degrees $(\Vert \Sigma(x_1)\Vert _F,\Vert \Sigma(x_2)\Vert _F,\cdots,\Vert \Sigma(x_M)\Vert _F)$ and the ground truth distribution $(\eta_1,\eta_2,\cdots,\eta_M)$ of the $M$ training instances. 
Unlike POEs \cite{ref25}, such operation can optimize the prediction to be as close as possible to the human ratings rather than a standard normal distribution,
\begin{align}
	\mathcal{L}_{Dis} = \sum_{m=1}^M {\eta_m \cdot (\log(\eta_m)-\log(\Vert\Sigma(x_m)\Vert_F))},	
\end{align}
where $M$ is the total number of training instances.

It has been generally accepted in machine learning that data augmentation can improve the robustness of models. 
We think our uncertainty modeling can be considered as a specific form of data argumentation, because its process is similar to the feature-level data augmentation \cite{ref24,ref47}. 
Firstly, we build up a Gaussian distribution in the high-dimensional psychological scale space according to the human ratings. 
Then, we randomly sample from these Gaussian distributions for pairwise comparisons. 
This process can be considered as disturbing a single feature point on the latent space, which is the feature level augmentation. 
As the disturbed features usually belong to the same class of the original feature, such augmentation is often applied to classification tasks \cite{ref24,ref47}. 
Our order learning just is a triple classification problem.

\subsection{Comparison of Distribution}
\label{3.3}
After modeling the FB uncertainty, an order should be established for these multi-dimensional Gaussian distributions on the phycological scale space. However, the conventional order learning cannot learn the order between distributions. 
Thurstone compares the observations of multiple subjects as the order of two stimuli on the psychological scale. 
To mimic this process, we design a uncertainty-oriented comparison module based on the Monte Carlo sampling.

To have a better order relation, we first constrain the Wasserstein distance between distributions on the psychological scale space. 
It allows that instances with similar scores have smaller distances between their distributions, while instances with significant different scores have larger distances. 
To this end, we apply a hinge loss to constrain the ordinal property of the psychological scale space and form a triplet for any three instances $(x_l,x_m,x_n )$ from the dataset, who have ground truth $(y_l,y_m,y_n )$ and corresponding feature distributions $(z_l,z_m,z_n )$,
\vspace{10pt}
\begin{figure}[t]
	\centering
	\includegraphics[width=0.8\linewidth]{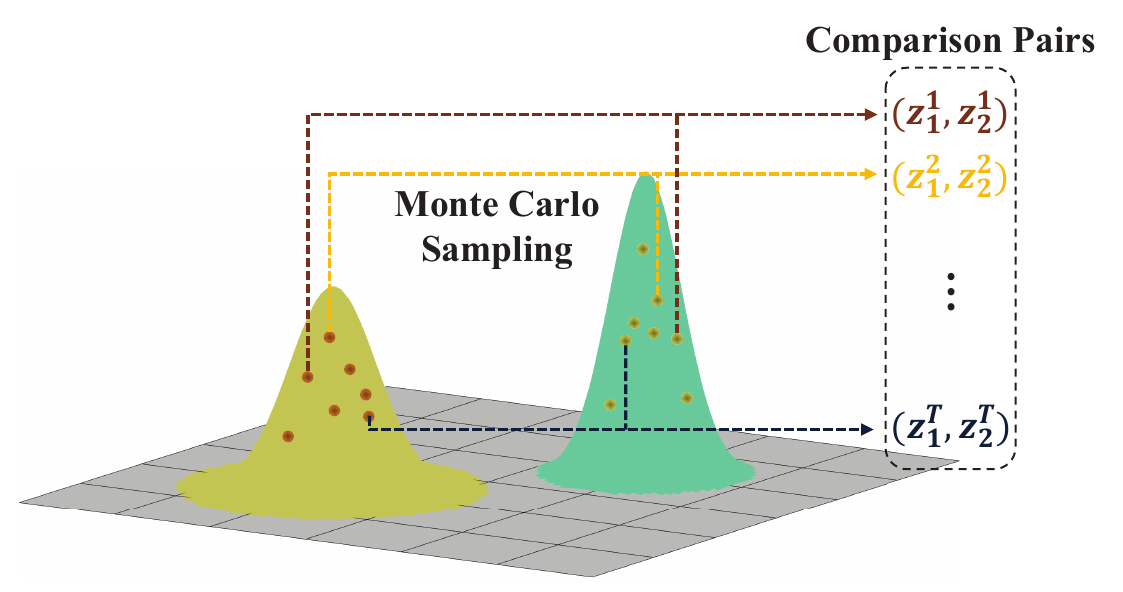}
	\caption{Monte Carlo sampling of our distribution comparison module.}
	\vspace{-0.3cm}
	\label{fig:sampling}
\end{figure}
\begin{align}
	\mathcal{L}_{Ord} = \frac{1}{\left\vert S \right\vert}\sum_{(l,m,n) \in S} \!{max(0,d(z_l,z_m)+\tau - d(z_l,z_n)),}
\end{align}
where $S=\{(l,m,n) \,\vert \,\left\vert y_l-y_m\right\vert < \left\vert y_l-y_n\right\vert \}$ and $\tau$ is the margin. 
$d(\cdot,\cdot)$ denotes the Wasserstein distance between two Gaussian distributions,
\begin{align} 
	d(z_1,z_2)^2= \sum_{j=1}^D{(\mu_1^j-\mu_2^j)^2+(\sigma_1^j -\sigma_2^j)^2},
\end{align}
where $\mu_1^j$, $\mu _2^j$, $\sigma _1^j$ and $\sigma _2^j$ are the $j$-th dimension of $\mu _1$, $\mu _2$, $diag(\Sigma _1)$ and $diag(\Sigma _2 )$ respectively. $D$ is the dimensionality of the vector.
The construction procedure of triplets can be found in Appendix A.

Afterwards, we apply $T$ times Monte Carlo sampling on the distribution of instance $x_i$, which is analogous to the observations of multiple subjects on a stimulus. 
To make network be backpropagated, the random sampling and forward propagation must be separated. 
Thus, we apply the reparameterization sampling method \cite{ref26} to get the $t$-th sampling $z_i^{(t)}$ from distribution $z_i$, 
\begin{align}
	z_i^{(t)} = \mu(x_i) + diag(\sqrt{\Sigma(x_i)})\cdot \varepsilon^{(t)}, \,\varepsilon^{(t)} \sim \mathcal{N}(0,I), 
\end{align}
where $\mathcal{N}(0,I)$ denotes the multi-dimensional Gaussian distribution with zero mean and identity covariance matrix $I$.
The sampling process is shown in Fig. \ref{fig:sampling}.

The comparator $f (\cdot,\cdot)$ in conventional order learning is applied to learn the order between two sampling feature points. 
The relative relation $Y^\prime$ between two distributions of $x_1$ and $x_2$ is obtained by calculating the mean of $T$ comparisons, 
\begin{align}
	Y^\prime = \frac{1}{T}\sum_{t=1}^T{f(z_1^{(t)},z_2^{(t)})}.
\end{align}

A cross-entropy loss $\mathcal{L}_{Cls}$ for triple classification is used to optimize the comparator $f (\cdot,\cdot)$,
\begin{align}
	\mathcal{L}_{Cls} = -\log \frac{\exp (Y_c^\prime)}{\sum_{r=1}^3 \exp(Y_r^\prime)},
\end{align}
where $Y_c^\prime$ and $Y_r^\prime$ denote the $c$-th and $r$-th dimensions of the output vector $Y^\prime $, $c$ is the dimension where the ground truth is.

Thus, the entire loss of our UOL is
\begin{align}
	\mathcal{L} = \mathcal{L}_{Cls}+\alpha \mathcal{L}_{Ord} + \beta\mathcal{L}_{Dis},
\end{align}
where $\alpha$ and $\beta$ are weights to control the contribution of each loss function.

\begin{figure*}[t]
	\centering
	\includegraphics[width=0.85\linewidth]{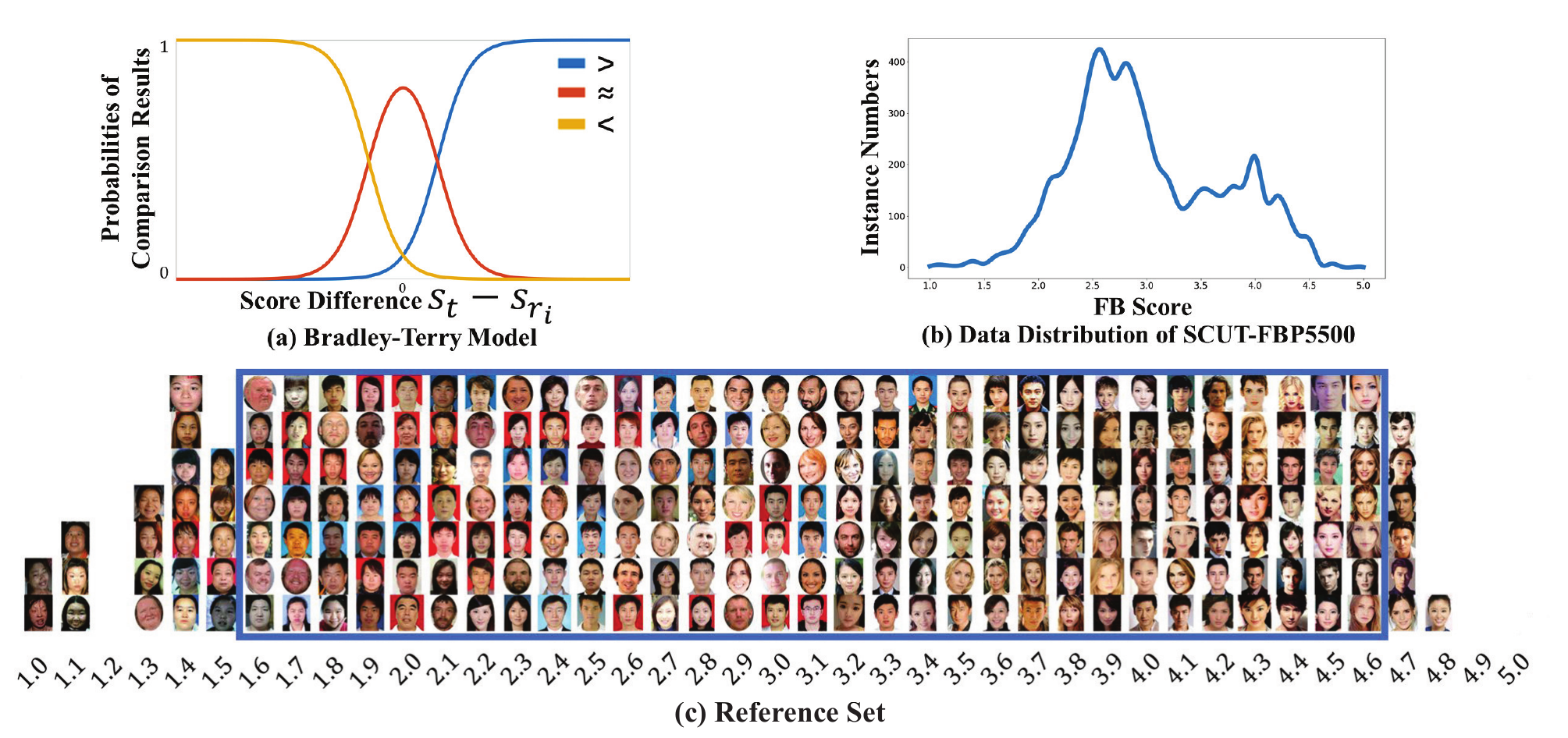}
	\caption{Illustration of Bradley-Terry (BT) estimation module. (a) shows the probability distribution of BT model. (b) shows the data distribution of SCUT-FBP5500 dataset. (c) lists the reference set, all reference images with precise scores are going to be compared with the input image to estimate its FB score. The range marked by blue box is the reference set selected by MC rules \cite{ref29}, which cannot cover the entire range.}
	\vspace{-0.3cm}
	\label{fig:BT}
\end{figure*}
\subsection{FB Score Estimation by Bradley-Terry Model}
\label{3.5}
After establishing the order of samples, network cannot predict the FB score yet because the order is independent on the reference base. 
Conventional order learning \cite{ref29} compares the input face image with a set of reference images whose labels cover the entire range of ages. 
These comparisons will find the most similar references to the input. 
Their precise labels will determine the label of input. 
This method is known as the maximum consistency (MC) rule \cite{ref29}, which requires the reference set must be balanced (the number of reference images must be the same for each interval) and continuous (no discontinue interval throughout the entire range). 

\label{sect:BT}

However, most FBP datasets are unbalanced. 
Figure \ref{fig:BT} (b) shows an example, the data distribution of SCUT-FBP5500.
Thus, MC rule can only cover the range of 1.6$ \sim $4.5, the blue box in Fig. \ref{fig:BT} (c). 
To address this problem, we propose a score estimation method based on Bradley-Terry model. 
Specifically, an input with unknown score $s_t$ is compared with a reference image with known score $s_{r_i}$. 
Bradley-Terry model tries to estimate the best $s_t$, and then models the possible order result $Y$ and score difference $(s_t - s_{r_i})$ as the following probability distribution,
\begin{align}
	\begin{split}
		& P(Y\le y_j,s_{r_i};s_t) = \frac{e^ {\delta_{y_j}+(ks_t-ks_{r_i})}}{1+e^ {\delta_{y_j}+(ks_t-ks_{r_i})}},\\
		&\quad s_{r_i} \in S_{ref}, \quad y_j \in\{0,1,2\},
	\end{split}
\end{align}
where 0, 1 and 2 represent the ``$<$", ``$\approx$" and ``$>$" relations. $\delta_0 = -\delta_1 = -\delta$, $\delta_2 = + \infty$, $\delta$ is a positive parameter to control the probability of ``$\approx$". $k$ is a scaling parameter to determine the change rate of probability. $S_{ref}$ denotes the set of all scores in the reference set.

Suppose $n$ images exist in the reference set and their ground truth scores are $\{s_1, s_2, \dots , s_n\}, s_i \in S_{ref}$. 
We apply the optimized comparator $f (\cdot,\cdot)$ to predict the order between the input and each reference image, which results in $\{R_1, R_2, \dots, R_n\}$, $R_i \in \{0,1,2\}$, then maximize the likelihood function, 
\begin{align}
	L(s_t) = \prod_{i=1}^n {P(Y=R_i, s_i; s_t)}.
\end{align}

Finally, the FB score $s_t$ of input image can be obtained.

The advantage of using maximum likelihood estimation is that FB scores can be estimated in the entire range by partial comparison results. 
Therefore, our UOL can work on unbalanced and discontinuous reference set. 
Obviously, the more complete the reference set is, the better performance our UOL achieves. 
Figure \ref{fig:BT}(c) shows the reference set for our UOL based on SCUT-FBP5500 dataset.

\section{Experiment}
\subsection{Implementation Details}
We use a pretrained VGG16 on ImageNet as the backbone of our method, and optimize it by Adam optimizer with a batch size of 32. 
The learning rate is $1e-4$ at the beginning and a Cosine Annealing scheduler with the minimal learning rate $1e-6$ is applied. 
We set the training epoch to 100 for all 5-fold cross validation. For data preprocessing, all the facial images ($350\times350$) are resized to $256\times256$ firstly. 
Then a $224\times224$ center croping and a random horizontal flipping are performed, followed by per-pixel rescale to $0 \sim 1$ and mean value subtraction. 
The hyperparameters $\alpha$ and $\beta$ in the loss function are $1e-4$ and $1e-3$, respectively.

We discretize the FB scores of training set at intervals of 0.1, and select $min(n_i,10)$ images in each score interval from training set as the reference images to estimate the final score, where $n_i$ denotes the total number of training data in the $i$-th score interval.

\subsection{Datasets and Evaluation Metrics}

\subsubsection{Datasets}
\indent \textbf{SCUT-FBP5500} \cite{ref28} has 5500 frontal facial images, which was scored by 60 volunteers among the range of 1$ \sim $5. 
The data consist of male/female and Asian/Caucasian faces, and have diverse annotation information (facial feature annotation, ratings and mean rating for each face by different volunteers). 
In this paper, we use the mean score and corresponding variance of different volunteers' ratings to train and test our model.

\textbf{Hot-Or-Not} \cite{ref16} contains 2056 frontal female facial images aged 18-40 without constraint on race, lighting, pose or expression. 
The data was scored by 30 volunteers among the range of -3$ \sim $3.

\textbf{MEBeauty} \cite{ref22} includes 1300 females and 1250 males facial images. 
Each gender group is divided into six racial categories: Black, Asian, Caucasian, Hispanic, Indian and Middle Eastern. 
The data was scored by 300 volunteers among the range of 1$ \sim $10.

\textbf{Color FERET} \cite{ref38} is a dataset for face recognition. 
It contains 11,338 color images of size 512×768 pixels captured in a semi-controlled environment with 13 different poses from 994 subjects. 
In this paper, 671 frontal images are selected to validate the generalization capability of our method.

\textbf{MIFS} (Makeup Induced Face Spoofing) \cite{ref8} is a facial image dataset collected from YouTube videos of makeup impersonations, consisting of 107 makeup transformations. 
Each subject has two images with and without makeup. 
In real life, people usually believe faces who wear makeup are more attractive than those who do not. 
Therefore, we select facial image of each subject with and without makeup to evaluate the discrimination of UOL and competing methods.

In this work, all methods are only trained on the training set of SCUT-FBP5500, and evaluated without any fine-tuning on all five datasets, in which Hot-Or-Not, MEBeauty, Color FERET, and MIFS are unseen datasets for all methods during test. Such setting is for testing the generalization ability of UOL.
\subsubsection{Evaluation Metrics}
\indent To test the effectiveness of UOL, we follow the evaluation metrics in \cite{ref31,ref52,ref42,ref30,ref53,ref63}: mean absolute error (MAE) and root mean square error (RMSE). 
MAE measures the mean of absolute errors between the predictions and ground truth. 
RMSE measures the deviation between the predictions and ground truth. 
However, lower MAE and RMSE do not guarantee a better correlation of predictions and ground truth. 

As SCUT-FBP5500, Hot-Or-Not and MEBeauty have varied ranges of FB scores rated by different people, MAE and RMSE are infeasible to evaluate the generalization ability of a method. 
The Pearson correlation coefficient (PC) \cite{ref4} is a well-accepted metric for the evaluation of FB in psychological research.
PC quantifies the degree of interdependence of prediction and ground truth. 
Thus, we employ PC as the metric to measure the generalization ability of a model trained on SCUT-FBP5500.

For MIFS dataset, we apply the accuracy rate (ACC) as the metric, which measures if the estimation of a face with makeup by a model is higher than that of the face without makeup.
\begin{table}[t]
	\makeatletter\def\@captype{table}\makeatother\caption{Performance comparison on SCUT-FBP5500. The results of competing methods are from their papers.}
	\centering      
	\begin{tabular}{lccc} 
		\hline\noalign{\smallskip}
		Methods                   & PC $\uparrow$ & MAE $\downarrow$ & RMSE $\downarrow$ \\
		\hline
		\noalign{\smallskip}
		$R^3$CNN \cite{ref30}     & 0.9142        & 0.2120           & 0.2800       \\
		CRNet \cite{ref52}        & 0.8869        & 0.2397           & 0.3186       \\
		AaNet \cite{ref31}        & 0.9055        & 0.2236           & 0.2954       \\
		ComboLoss \cite{ref53}    & 0.9199        & 0.2050           & 0.2704       \\
		Co-Attention \cite{ref42} & {\bf 0.9260}  & 0.2020           & 0.2660       \\
		CNN-ER \cite{ref63}       & 0.9250        & 0.2009           & 0.2650       \\
		{\bf UOL}                 & 0.9240        &{\bf 0.1975}      &{\bf 0.2633}  \\
		\hline
	\end{tabular}
	\vspace{-4pt}
	\label{table:table1}
\end{table}
\begin{table}[t]
		\makeatletter\def\@captype{table}\makeatother\caption{Performance of different models on the Hot-Or-Not.}
		\centering  
		\begin{tabular}{lccc}
			\hline
			\noalign{\smallskip}
			Methods & PC $\uparrow$ & MAE $\downarrow$ & RMSE $\downarrow$  \\
			\hline
			\noalign{\smallskip}
			$R^3$CNN \cite{ref30}     & 0.3555      & 0.5741      & 0.7140         \\
			CRNet \cite{ref52}        & 0.3250      & 0.5811      & 0.7294         \\
			AaNet \cite{ref31}        & 0.2893      & 0.5923      & 0.7399         \\
			ComboLoss \cite{ref53}    & 0.3329      & 0.6154      & 0.7677         \\
			Co-attention \cite{ref42} & 0.2697      & 0.5613      & 0.7080         \\
			CNN-ER \cite{ref63}       & 0.3513      & \bf{0.5269} & \bf{0.6653}    \\
			\bf{UOL}                  & \bf{0.4073} & 0.5410      & 0.6779         \\
			\hline
		\end{tabular}
		\label{table:table2}
\end{table}

\subsection{Comparison with SOTA Methods}
To verify the performance and generalization ability of UOL, we compare it with the state-of-the-art methods $R^3$CNN \cite{ref30}, CRNet \cite{ref52}, Co-attention \cite{ref42}, AaNet \cite{ref31},  ComboLoss \cite{ref53} and CNN-ER \cite{ref63}.
\subsubsection{Performance Evaluation on SCUT-FBP5500}
\indent We firstly test all methods on the large-scale dataset SCUT-FBP5500, and report the result in Table~\ref{table:table1}. 
One can see that our UOL achieves the best on both MAE and RMSE, but slightly worse than Co-attention and CNN-ER on PC, which demonstrates that UOL has reached the state-of-the-art performance on SCUT-FBP5500.

\subsubsection{Generalization Capability Evaluation}
\indent Our method aims to improve the generalization ability of the model by mimicking the human cognition. 
To this end, we train all methods on SCUT-FBP5500, and then test them on unseen datasets (Hot-Or-Not, MEBeauty, Color FERET, MIFS) which are not used to fine tune models. 
The competing methods are strictly implemented according to their open codes and papers.

\textbf{(1) Experiments on Datasets with Human Ratings}

We normalize scores of Hot-Or-Not and MEBeauty to the range 1$ \sim $5 and compute PC with the estimated scores by each method. 
Tables~\ref{table:table2} and ~\ref{table:table3} show that the PC of UOL is considerably higher than those of other methods, which indicates that UOL has better generalization ability.
The evaluations of MAE and RMSE show that UOL underperforms CNN-ER on Hot-Or-Not by $1\%$ $\sim$ $2\%$, but outperform it on MEBeauty by $5\%$.
One can see Co-attention performs worse than on SCUT-FBP5500, which shows it is sensitive to test data when FB standard shifts or image quality varies.

\textbf{(2) Experiments on Datasets without Human Ratings}

Color FERET and MIFS do not have human ratings. 
So we design two different experiments to evaluate the generalization ability of these methods.

\textbf{For Color FERET}, we select all 671 frontal face images as the test data, and apply UOL and six competing methods to give scores for each image for simulating human rating. 
After having all six ratings for each image, the lowest and highest ones are removed, the mean of the remaining ratings is considered as FB score. 
Afterwards, we do the same process on datasets with human ratings, and list the results in Table~\ref{table:table4}. UOL also achieves the highest PC.

\begin{table}[t]
		\makeatletter\def\@captype{table}\makeatother\caption{Performance of different models on the MEBeauty.}
		\centering  
		\begin{tabular}{lccc}
			\hline
			\noalign{\smallskip}
			Methods  & PC $\uparrow$ & MAE $\downarrow$ & RMSE $\downarrow$  \\
			\hline
			\noalign{\smallskip}
			$R^3$CNN \cite{ref30}     & 0.5039      & 0.5329      & 0.6691      \\
			CRNet \cite{ref52}        & 0.4380      & 0.5645      & 0.9019      \\
			AaNet \cite{ref31}        & 0.3746      & 0.6102      & 0.7548      \\
			ComboLoss \cite{ref53}    & 0.5078      & 0.5481      & 0.6888      \\
			Co-attention \cite{ref42} & 0.4976      & 0.5476      & 0.6907      \\
			CNN-ER \cite{ref63}       & 0.4911      & 0.5753      & 0.6973      \\
			\bf{UOL}                  & \bf{0.5532} & \bf{0.5230} & \bf{0.6489} \\
			\hline
		\end{tabular}
		\label{table:table3}
		\vspace{-4pt}
\end{table}
\begin{table}[t]
	\caption{Comparison of the generalization ability of seven models on Color FERET and the accuracies on MIFS.} 
	\centering
	\begin{tabular}{lcccccc} 
		\hline
		\noalign{\smallskip}
		Methods & {\bf PC $\uparrow$ (Color FERET)} & {\bf Acc(\%) $\uparrow$ (MIFS)}  \\
		\hline
		\noalign{\smallskip}
		$R^3$CNN \cite{ref30}     & 0.8265        & 73.91         \\
		CRNet \cite{ref52}        & 0.8564        & {\bf 96.15}   \\
		AaNet \cite{ref31}        & 0.7504        & 76.92         \\
		ComboLoss \cite{ref53}    & 0.9146        & 92.31         \\
		Co-Attention \cite{ref42} & 0.7067        & 76.92         \\
		CNN-ER \cite{ref63}       & 0.8490        & {\bf 96.15}   \\
		{\bf UOL}                 & {\bf 0.9266}  & 92.31         \\
		\hline
	\end{tabular}
	\label{table:table4}
\end{table}

\textbf{For MIFS}, facial images appear in pairs, in which one is with makeup and another one is not. 
We employ 7 volunteers to compare the image pairs, and clean the data according to the consistency of volunteers' results as follows:

\textbf{Step 1}: Manually select two frontal facial images of each face ID with and without makeup from MIFS, respectively. Group them as a pair.

\textbf{Step 2}: Show each pair to 7 volunteers. 
They vote the more beautiful image in the pair.

\textbf{Step 3}: Calculate the consistency of the volunteers' votes of each face ID, select pairs with higher consistency (more than 5 volunteers give the same vote) as the test data. The ground truth is the majority voting of volunteers. 

We apply each method on images in a pair. 
If the estimated score of image with makeup is higher than the one without makeup, the comparison is correct, otherwise incorrect.
Table~\ref{table:table4} shows UOL is the second best. 
After carefully examining the results, we find the two pairs misestimated by UOL are also misestimated by other four methods. 
It indicates that some unknown FB features have not been explored by these methods. 
CRNet just misestimates a pair, the best in this experiment, but performs worse than its upgrade version, ComboLoss, on other experiments.  

All above results demonstrate the generalization ability of UOL outperforms the competing FBP algorithms.

\begin{table*}[t]
	\caption{The effectiveness evaluation of uncertainty modeling and order learning on SCUT-FBP5500 ({\bf SCUT}), Hot-Or-Not ({\bf HON}) and MEBeauty ({\bf MEB)}.}   
	\centering  
	\begin{tabular}{lcccccc} 
		\hline
		\noalign{\smallskip}
		\multirow{2}{*}{Methods} & \multicolumn{3}{c}{\bf SCUT} & {\bf HON} & {\bf MEB} \\
		\cline{2-6}
		\noalign{\smallskip}
		& PC$\uparrow$ & MAE$\downarrow$ & RMSE $\downarrow$ & PC$\uparrow$ & PC$\uparrow$\\
		\hline
		\noalign{\smallskip}
		VGG16(Regression)                        & 0.9044      & 0.2248      & 0.2973      & 0.3675      & 0.5122       \\
		VGG16(Regression) + LDL                  & 0.9076      & 0.2214      & 0.2909      & 0.3228      & 0.5338       \\
		VGG16(Regression) + Uncertainty Modeling & 0.9080      & 0.2218      & 0.2920      & 0.3895      & 0.5367       \\
		VGG16(Order Learning)                    & 0.9198      & 0.2025      & 0.2683      & 0.3958      & 0.5442       \\
		{\bf UOL}                                & \bf{0.9240} & \bf{0.1975} & \bf{0.2633} & \bf{0.4073} & \bf{0.5532} \\
		\hline
	\end{tabular}
	\label{table:table6}
\end{table*}
\begin{table*}[t]
	\setlength{\belowcaptionskip}{0.1cm}
	\makeatletter\def\@captype{table}\makeatother\caption{The effectiveness evaluation of three loss functions on Hot-Or-Not and MEBeauty.}
	\centering
	\begin{tabular}{ccccccccccccc}
		\hline
		\noalign{\smallskip}
		\multirow{2}{*}{CE} & & \multirow{2}{*}{Hinge} & & \multirow{2}{*}{KL} & &  \multicolumn{3}{c}{\bf Hot-Or-Not}  & & \multicolumn{3}{c}{\bf MEBeauty}  \\
		\cline{7-13} 
		\noalign{\smallskip}
		& & & & & & PC $\uparrow$ & MAE $\downarrow$ & RMSE $\downarrow$ & & PC $\uparrow$ & MAE $\downarrow$ & RMSE $\downarrow$  \\
		\hline
		\noalign{\smallskip}
		\checkmark & &            & &            & & 0.4007 & 0.5419 & 0.6844 & & 0.5457 & 0.5252 & 0.6543 \\
		\checkmark & & \checkmark & &            & & 0.4036 & 0.5410 & 0.6793 & & 0.5463 & 0.5231 & 0.6508 \\
		\checkmark & &            & & \checkmark & & 0.4036 & 0.5426 & 0.6792 & & 0.5397 & 0.5290 & 0.6592 \\
		\checkmark & & \checkmark & & \checkmark & & \bf{0.4073} & \bf{0.5410} & \bf{0.6779} & & \bf{0.5532} & \bf{0.5230} & \bf{0.6489} \\
		\hline\hfill
	\end{tabular}
	\vspace{-12pt}
	\label{table:table8}
\end{table*}
\subsection{Ablation Studies}
\subsubsection{Effectiveness of order learning and uncertainty modeling}
\indent To validate the effectiveness of order learning and uncertainty modeling respectively, we conduct ablation studies on three datasets with FB scores. 
It is worth noting that we do not separately evaluate Bradley-Terry module, because UOL cannot estimate scores covering whole range without it. 
All versions are trained on the SCUT-FBP5500. 
The backbone is VGG16. 
The results under different settings are listed in Table~\ref{table:table6}. 
One can see that order learning contributes a significant performance gain, because it is more consistent with human cognition to order patterns than regression approaches. 
Uncertainty modeling also boosts the performance of a regression model with a marginal gain. 
Their integration, UOL, can further boost the performance. 
These results demonstrate that order is a very valuable invariant in FB, and uncertainty modeling is more feasible for order learning (a classification model) than a regression model.

Label distribution learning (LDL) could be also considered an uncertainty modeling. 
We then apply LDL to the backbone and report the results in Table~\ref{table:table6}. 
It can be seen that LDL achieves a marginal gain on SCUT-FBP5500 and MEBeauty, but performs worse on Hot-Or-Not. 
The possible reason is that SCUT-FBP5500 and MEBeauty have similar data distributions, but Hot-Or-Not has different distribution. LDL essentially learns the mapping from fixed points on the latent space to certain distributions and is inclined to overfit the label distribution. 
Please note that order learning has difficulty in using label distribution because its form is not comparable to learn order.
\subsubsection{Effectiveness of three losses}
\indent UOL employs three loss functions that play different roles.
CE Loss aims at training the comparator in Fig. \ref{fig:framework} for estimating the relative order between instances. CE Loss is indispensable to our UOL. 
Hinge Loss constrains the distance between the modeled distributions of instances on the latent space, which can improve the representation of order relation and further boost the order estimation.
KL Loss constrains the consistency between the modeled distribution of instances and the variance of human ratings for instances, which aims at a more accurate distance metric in Hinge Loss.

We also evaluate the effectiveness of them and report the results in Table~\ref{table:table8}. 
We can see that Hinge Loss helps CE Loss achieve better performance, KL Loss further boosts the performance when CE Loss and Hinge Loss work together. But CE $+$ KL Losses degrade the performance because KL Loss cannot directly help CE Loss without Hinge Loss.

\section{Conclusion}
In this paper, we propose a novel Uncertainty-oriented Order Learning for facial beauty prediction. 
UOL enables order learning to learn the relative relations of uncertain data by a distribution comparison module, in which order learning addresses the inconsistency of FB standards between datasets, and uncertainty modeling tackles the inconsistency of human cognition of FB. 
In addition, we introduce the Bradley-Terry model into order learning to relax the restriction that the reference set must be continuous and balanced. 
Extensive experiments demonstrate that our method outperforms state-of-the-art methods on SCUT-FBP5500 in terms of FB score prediction, and better generalization on other datasets.
However, the improper use of FBP models might result in an unethical impact. Devising better data forensics approaches
could be countermeasures. 
In the future work, we will explore the impact of face attributes on the UOL.

\bibliography{reference}
\bibliographystyle{IEEETran}
%

%
%
%




\clearpage
\appendices
\section{Triplet Construction for Hinge Loss}
\label{apd1}
To model the order on psychological scale space described in \textbf{Sec. 3.3}, we construct a triplet of instances $(x_l,x_m,x_n)$ for the hinge loss, which ensures that their FB scores $(y_l,y_m,y_n)$ meet $|y_l-y_m |<|y_l-y_n |$. 
Theoretically, selecting instances can be random, it makes training time consuming. Therefore, we define a hard triplet, whose difference between $|y_l-y_m |$ and $|y_l-y_n |$ is small, and construct $M$ hard triplets from a batch of training data with $M$ instances. 
We take the $i\verb|-|th$ instance as the anchor $l$, then set the instance with index $(i+1)\ {mod}\ M$  as the second element $m$. Finally, we select an instance $n$ from the remaining $M-2$ instances as the third element, which satisfies:
\begin{align}
	\begin{split}
		&\mathop{\arg\min}\limits_{n \neq l,m}\left\vert \left\vert y_l-y_m\right\vert   - \left\vert y_l-y_n\right\vert \right\vert, \  \\&{\bf where} \  \left\vert y_l-y_m\right\vert \neq \left\vert y_l-y_n\right\vert . 
	\end{split}
\end{align}

Algorithm \ref{al1} describes the strategy of this hard triplets selection.

\IncMargin{1em}
\begin{algorithm}
	\small \SetKwData{Left}{left}\SetKwData{This}{this}\SetKwData{Up}{up} \SetKwFunction{Union}{Union}\SetKwFunction{FindCompress}{FindCompress} \SetKwInOut{Input}{Input}\SetKwInOut{Output}{Output}
	\caption{Hard Triplets Selection}
	\label{al1}
	\Input{FB scores of $M$ (batch size) instances $y=\{y_1, y_2, \dots, y_M\}$.}
	\Output{Triplets of index, originized in batch form, ($l=\{l_1, l_2, \dots, l_M\}$, $m=\{m_1, m_2, \dots, m_M\}$, $n=\{n_1, n_2, \dots, n_M\}$).\\
	}
	\BlankLine
	\For{$i\leftarrow 1$ \KwTo $M$}{
		$l_i \leftarrow i$\;
		$m_i \leftarrow (i+1) \mod M$\;
		$dis_{12} \leftarrow \left\vert y_{l_i} - y_{m_i} \right\vert$\;
		$sub = +\infty$\;
		\For{$j\leftarrow 1$ \KwTo $M$}{
			\If{$j \neq  l_i$ {\bf and} $j \neq m_i$}{
				$dis_{13} \leftarrow \left\vert y_{l_i} - y_{j} \right\vert$\;
				$tmp \leftarrow \left\vert dis_{12} - dis_{13} \right\vert$\;
				\If{$tmp < sub \ {\bf and}\ tmp \neq 0$}{
					$sub \leftarrow tmp$\;
					$n_i \leftarrow j$\;
				}
			}
			
		}
	}
\end{algorithm}
\DecMargin{1em} 

\newpage
\section{Pair Construction for Pairwise Comparison}
Assume FB scores are in the range of $\left[ 1, H\right] $. 
In order learning, we define a pair of two instances, who meet $|y_i-y_j |< \theta$, as a ``$\approx$" pair, where $\theta \ll H$. If instances are selected randomly in each batch, the number of ``$ \approx $" pairs will be much smaller than ``$>$" pairs and ``$<$" pairs. 
Such data imbalance makes the trained comparator in \textbf{Sec. 3.3} overfit the ``$>$" and ``$<$" pairs. 
Meanwhile, random selection may let an instance frequently appear in a batch. 
It also makes networks overfit this instance. 
To tackle these problems, we design a balanced pairs selection strategy, which ensures the proportion of ``$\approx$" pairs in a batch is close to $\frac{1}{3}$, and all instances appear almost equally in a batch.  
Algorithm \ref{al2} describes our balanced pairs selection strategy.

\IncMargin{1em}
\begin{algorithm}
	\small 
	\SetKwData{Left}{left}\SetKwData{This}{this}\SetKwData{Up}{up} \SetKwFunction{Union}{Union}\SetKwFunction{FindCompress}{FindCompress} \SetKwInOut{Input}{Input}\SetKwInOut{Output}{Output}
	\caption{Balanced Pairs Selection}
	\label{al2}
	\Input{FB scores of $M$ (batch size) instances $y=\{y_1, y_2, \dots, y_M\}$; adjacency list $flag=\{flag[1], flag[2], \dots, flag[M]\}$, where $flag[i]$ is an empty list denotes which instances have been paired with the $i$-th instance; limitation of every instance can be selected $N$; threshold $\theta$.}
	\Output{Pairs of index $p$, and corresponding order relationships $Y$.}
	\BlankLine
	
	$p \leftarrow \{\ \}$\;
	$Y \leftarrow \{\ \}$\;
	\For{$i\leftarrow 1$ \KwTo $M$}{
		$candidates \leftarrow \{1, 2, \dots, M\}$\; \tcp*[h]{$candidates$ is a list denotes which instances can be selected.}\\
		{DELETE}$(candidates, i)$\; \tcp*[h]{DELETE denotes deleting an element from a list.}\\
		\For{$j \leftarrow 0$ \KwTo {LEN}$(flag[i])$}{
			DELETE$(candidates, flag[i][j])$\;
		} 
		\For{$j \leftarrow 1$ \KwTo $M$}{
			\If{$flag[i] > N$}{
				DELETE$(candidates, j)$\;
			}
			
		}
		\While{$flag[i] < N$ {\bf and} $candidates \neq \{ \ \}$}{
			$sim \leftarrow  0$\;
			\For{$j \leftarrow 1$ \KwTo LEN$(candidates)$}{
				$dif \leftarrow \left\vert y_i - candidates[j] \right\vert $\;
				\If{$dif < \theta$}{
					$sim \leftarrow sim + 1$\;\tcp*[h]{$sim$ is the number of "$\approx$" pairs in $candidates$.}\\
				}	
			}
			$unsim \leftarrow {LEN}(candidates) - sim$\;
			$prob = \{\ \}$\; \tcp*[h]{$prob$ is the list denotes probability of every instance to be selected.}\\
			\For{$j \leftarrow 1$ \KwTo LEN$(candidates)$}{
				$dif \leftarrow \left\vert y_i - candidates[j] \right\vert $\;
				\eIf{$dif < \theta$}{
					INSERT($prob$,$\frac{1}{3 * sim}$)\;\tcp*[h]{INSERT denotes appending an element to a list.}\\
				}
				{
					INSERT($prob$,$\frac{2}{3 * unsim}$)\;
				}
			}
			$r \leftarrow$ RANDOM\_CHOICE\_BY\_PROB$(candidates, prob)$\; \tcp*[h]{select an instance's index $r$ from $candidates$ by their probabilities $prob$, we implement this by using numpy.}\\
			INSERT$(flag[i], r)$\;
			INSERT$(flag[r], i)$\;
			INSERT$(p,\{i,r\})$\;
			$order \leftarrow$ GENLABEL$(\left\vert y_i - y_r \right\vert,\theta)$\;\tcp*[h]{GENLABEL denotes generating the order label for a pair by their ground truth and threshold.}\\
			INSERT$(Y, order)$\;
			DELETE$(candidates, r)$\;
			
		}
		
	}	
\end{algorithm}
\DecMargin{1em} 

\end{document}